**Longevity Associated Geometry Identified in Satellite Images: Sidewalks, Driveways and Hiking Trails**

Joshua J. Levy, B.A.[1,2,3,*], Rebecca M. Lebeaux, B.S.[1,2], Anne G. Hoen, Ph.D.[1,2,4], Brock C. Christensen, Ph.D.[1], Louis J. Vaickus, MD Ph.D.[3], Todd A. MacKenzie, Ph.D.[1,4,5]

1. Department of Epidemiology, Geisel School of Medicine at Dartmouth, Lebanon, New Hampshire, USA
2. Program in Quantitative Biomedical Sciences, Geisel School of Medicine at Dartmouth, Lebanon, New Hampshire, USA
3. Department of Pathology, Dartmouth Hitchcock Medical Center
4. Department of Biomedical Data Science, Geisel School of Medicine at Dartmouth, Lebanon, New Hampshire, USA
5. The Dartmouth Institute for Health Policy and Clinical Practice, Geisel School of Medicine at Dartmouth, Lebanon, New Hampshire, USA

*To whom correspondence should be addressed: joshua.j.levy.gr@dartmouth.edu

Word Count: 3000 words

**Key Points**

**Question:** What is the relationship between mortality and satellite images as elucidated through the use of Convolutional Neural Networks?

**Findings:** Direct prediction of mortality using a deep learning model across a cross-sectional study of 430 US counties identified key features in the environment (i.e. sidewalks, driveways and hiking trails) associated with lower mortality.

**Meaning:** Deep learning tools that utilize satellite imagery have the potential to identify public health interventional targets.

## Abstract

**Importance:** Following a century of increase, life expectancy in the United States has stagnated and begun to decline in recent decades. Using satellite images and street view images prior work has demonstrated associations of the built environment with income, education, access to care and health factors such as obesity. However, assessment of learned image feature relationships with variation in crude mortality rate across the United States has been lacking.

**Objective:** Investigate prediction of county-level mortality rates in the U.S. using satellite images.

**Design:** Satellite images were extracted with the Google Static Maps application programming interface for 430 counties representing approximately 68.9% of the US population. A convolutional neural network was trained using crude mortality rates for each county in 2015 to predict mortality. Learned image features were interpreted using Shapley Additive Feature Explanations, clustered, and compared to mortality and its associated covariate predictors.

**Main Outcomes and Measures:** County mortality was predicted using satellite images.

**Results:** Predicted mortality from satellite images in a held-out test set of counties was strongly correlated to the true crude mortality rate (Pearson r=0.72). Learned image features were clustered, and we identified 10 clusters that were associated with education, income, geographical region, race and age.

**Conclusion and Relevance:** The application of deep learning techniques to remotely-sensed features of the built environment can serve as a useful predictor of mortality in the United States. Tools that are able to identify image features associated with health-related outcomes can inform targeted public health interventions.

# Introduction

Life expectancy in the United States has increased dramatically over the past century from 48 years in 1900 to 80 years in 2019. However, the United States has experienced a drop in longevity over the past decade and now ranks 43rd in the world [1–5]. Within the United States, crude mortality rates vary by more than 40%. Factors observed to be associated with mortality rates within the United States include disparities in socioeconomic status[6,7] and health insurance coverage,[8] as well as obesity, smoking[9,10] and drug use/opioid abuse[11].

Prior studies have attempted to infer characteristics of the underlying communities by characterizing land use using satellite imagery[12–14]. Recent research has leveraged deep learning approaches to link the built environment to obesity[15,16], socioeconomic status[17], poverty[18,19], and other demographic factors by integrating lower level image features into higher order abstractions to make predictions [20,21]. In addition, the potential for deep learning-based image analysis to characterize a broad range of health exposures was recently reviewed [22]. However, there is limited work that examines the relationship of mortality with the built environment in the United States at a large scale.

The identification of information in satellite images associated with mortality predictions could potentially unlock previously unknown, yet related geographic and structural community characteristics. Further, we propose that longitudinal image monitoring can be used to forecast increases in mortality rate and help determine optimal intervention strategies that complement county/state planning and development efforts.

We hypothesized that county-scale level mortality rates could be predicted from satellite images. The goals of this study were to determine whether the analysis of satellite images by

Convolutional Neural Networks (CNN) could be used to predict county-level mortality rates and uncover salient satellite image features associated with mortality. We also sought to determine if image features are related with county-level measures of income, education, age, sex, race and ethnicity. The main objective of this study was to highlight a proof-of-concept deep learning application that presents strong baseline performance, enabling future work that can evaluate the potential for applying learnt features to the design of public health interventions in the built environment.

## Methods

### Data Collection

To train our deep learning model, we used publicly available county-level mortality data. We used the CDC Wonder database to collect all available death data from 2015 for residents of the 50 United States and the District of Columbia, and matched death data with 2015 county Census population data from the Bureau of Economic Analysis, USDA ERS databases[23–25], and the Surveillance, Epidemiology, and Ends Results (SEER) Program[26]. Crude mortality rates were calculated as the number of deaths in the county divided by the population in the county. Additional county level covariate information from 2015 was collected on age, sex, Hispanic status, race, income, education, and region (Supplementary Table 1), though this covariate information was not used in our computer vision model.

We selected 430 counties from among 3,142 total United States counties in 2015 for inclusion in our study. To reduce the variance of the mortality estimate and to limit potential confounding from locale we selected the top 1000 most populous counties and split these counties into 13 bins containing similar numbers of counties rank ordered by mortality. We selected up to forty of the most populous counties in each bin. These selection criteria were developed to capture a greater proportion of total deaths and represent wider variation in mortality rates. 279 counties were placed into the training set (65%), 65 in the validation set (15%), and 86 in the test set

(20%). The training set was used to update the parameters of the trained model while the validation set was used to limit the model from overfitting to the training data (Figure 1a-b). The held-out test set represents an application of the modeling approach to unseen counties that had no role in the training of the model.

Satellite imagery data was collected using the Google Static Maps application programming interface (API) to build our deep learning pipeline, similar to Maharana and Nsoesie[15]. First, four schools from each county were randomly selected to serve as points of interest to sample nearby images. We used schools because they are typically placed in densely populated regions of the county. We obtained geographic coordinates for the schools in our study from the National Center for Education Statistics[27], and divided the 1 square mile area surrounding each school into an evenly spaced 7 images by 7 images grid (extracting 196 total images per county) (Figure 1c). The zoom level for each image was set to 17, and image dimensions were set to 400 pixels by 400 pixels to provide enough detail to make out street patterns and cover the space between the selected images of the grid. 84,280 satellite images were downloaded using the Google Static Maps API. Images were downloaded between July and September 2019, representing a collection of images acquired April 2018 to December 2018.

This study was exempt from institutional review board approval as we accessed previously collected data and could not identify individuals. Our study followed the Strengthening the Reporting of Observational Studies in Epidemiology (STROBE) guidelines for cross-sectional studies where appropriate[28].

**Deep Learning Model Training**

Deep neural networks have been used in a wide range of health-related applications[29,30]. Convolutional Neural Networks (CNN) slide filters across images to pick up on low level features such as edges or curves and then expand the visual field to pick up higher-order constructs

(Figure 1c) [31]. The particular variant of CNN that was used to predict mortality on the satellite images was ResNet-50 [32], pretrained on the ImageNet database to recognize over 1000 different objects [33], some of which may have features corresponding with our satellite image set. We performed transfer learning, which applies knowledge gained from ImageNet to initialize the parameters of our model. These model parameters were updated to minimize the divergence between the true and predicted mortality rates using the negative Poisson log likelihood; training the model with this objective allowed us to use each image to directly predict the mortality of its respective county. Training images were randomly cropped, rotated, resized, and flipped to improve the generalizability of the approach. Predictions averaged across the images for each county comprised the final prediction. After training our model on images from 279 counties, we evaluated the model on images from the remaining test counties and averaged the predicted mortality rates across each county to derive county mortality estimates. The validation set was used to terminate the learning process at 5 training iterations and identify the ideal learning rate for the model, 1e-4.

**Model Interpretation**

The output of an intermediate layer of the network was extracted from each image to form *embeddings* – reduced dimensional representations of the data. These *embeddings* (represented by 1000-dimensional vectors) could demonstrate how overall features of the images cluster and correspond with mortality or other demographics. These extracted image features were used to make UMAP [34] plots to identify clustering by images. We overlaid the actual images themselves for each of the points, then true mortality and other covariates such as education and aging. A model that has learned to recognized key image features would begin to cluster images with similar morphology. To examine specific associations between the pertinent image features (i.e. embeddings), we clustered image features using the Spectral Clustering approach [35] and found 10 clusters of images. We averaged the covariate information associated with each of the images across each grouping to yield characteristic covariate descriptors for each cluster. As a direct means for assessing the relationship between deep learning predicted mortality and the covariates, we regressed the image predicted mortality against each of these covariates.

A linear regression model to predict mortality using county level demographic characteristics was fit to data from both the training and validation sets, weighted by county population size, and evaluated on the test set as a comparison method versus the satellite image approach. We note that the goal of this study was to provide a benchmark for how well the CNN could predict mortality and to use covariates to help contextualize what the deep learning model is "seeing".

Shapley additive feature explanations (SHAP) [36] is an analytical technique that explains complex models using a simpler surrogate model for each testing instance. We applied SHAP to the images of the test counties to form pixel-wise associations with increases or reductions in mortality, hotspots in these images denote important mortality-associated objects that the model has learned; the technique was similarly applied to identify important covariate predictors for each county (Figure 1d). Finally, we convolved learned CNN image filters over select images to further demonstrate which features of the built environment were utilized by deep learning model.

## Results

The training, validation and test counties included 1,721,052 deaths per 217,938,597 individuals in the 2015 U.S. population, a crude mortality rate of 7.90 deaths per 1,000 U.S. residents (Supplementary Table 1).

Our deep learning model was able to accurately predict mortality, with a Pearson correlation coefficient of 0.72 ($R^2$=51%) between the predicted and true mortality (Figure 2a) on a held-out test set.

The linear regression model using county demographics also accurately predicted mortality; the $R^2$ was 90% (0.95 correlation) between covariates and the mortality rate (Figure 2b). We observed a reduction in 13 deaths per 100,000 individuals for an 1% increase in the proportion of those who attend college (regression coefficient $\beta = -12.6 \pm 0.8$), and increase of 214 deaths per 100,000 individuals for 5-year increases in average population age ($\beta = 2.1 \pm 0.1$); these were the most important predictors of county mortality (Figure 3a-c, Supplementary Table 2-5). Increased Hispanic ($\beta = -6.6 \pm 0.5$), female ($\beta = 42.5 \pm 6.4$), or Asian race proportions ($\beta = -7.8 \pm 1.0$), and living in the western United States ($\beta = -1.1 \pm 0.3$) as compared to the other regions, were found to be protective county-level factors against mortality (Figure 3a-b, Supplementary Tables 2-4).

We extracted 1000-dimensional vectors for each image from the second to last layer of the neural network to form embeddings that relate the images to each other by their features. We visualized these features using a 2-D UMAP plot. We grouped the images into ten clusters and found variation between clusters (Figure 3c-d, 4a). Weighted pairwise t-tests on mean differences between select covariates between all pairs of clusters demonstrated that image features also associate with covariate mortality predictors (Figures 4b, Supplementary Table 6). Images in cluster 7 were associated with the highest mortality rate of all of the clusters (mortality rate of 11 deaths per 1,000 individuals) and generally included counties from the southeastern United States (Supplementary Figure 5). Income and educational status were lower on average compared to the other clusters. We contrast cluster 7 with cluster 2, a small cluster with the third lowest mortality rates (6.6 deaths per 1,000 individuals) of the 10 clusters (Figure 4;

Supplementary Figures 6-8; Supplementary Table 6) and identified variation within cluster 5; we include a brief discussion in the supplementary materials. A univariable regression of the deep learning predicted mortality to each of the mortality predictors across the test set demonstrated that these image features are strongly associated with the mortality predictors (Supplementary Table 7).

Standardized regression coefficients from the covariate model and SHAP values over the test counties for the covariate model informed on the importance of each predictor (Figure 3a-b, Supplementary Tables 2-5). When we used SHAP to identify relevant features in satellite images across these test counties, we noticed that the model was able to associate common features associated with the socioeconomic status of that community to reductions in mortality. Generally, we were able to spot instances in which sidewalks, driveways, curved roads, hiking trails, baseball fields and light-colored roofs were associated with reductions in mortality (Figure 5). Conversely, we noticed instances where centerlines of large roads and shadows of buildings and trees were positively associated with mortality. To corroborate the evidence found using SHAP, we plotted small images representing patterns that the first convolutional layer had learned and then convolved each of these patterns with three select images (Supplementary Figures 1-4). Across three images from California, Virginia, and Georgia, filters such as 12, 29 and 43 focused on bright cues and were able to highlight driveways, sidewalks, walkways and baseball fields, while filters 13 and 46 were able to pick up on contours and nuanced shape-based patterns.

## Discussion

Here, we demonstrated the feasibility of using deep learning and satellite imagery to predict county mortality across the United States, extending methods and approaches from prior deep learning studies that linked the built environment and health-related factors. Furthermore, we established clusters of images that represented various demographic groups and interrogate these learned patterns to find additional corroborating evidence.

While the covariate model significantly outperformed our deep learning model and identified important predictors that corroborated with prior literature, the deep learning approach identified meaningful built features of the environment, representing a benchmark for performance from which to compare future applications[37,38].

We only sampled neighborhood characteristics found around schools. For instance, there is likely a greater number of sports fields and playgrounds surrounding schools than in random locations in the county. In many of the neighborhoods, regardless of mortality rates, there is likely more green space surrounding schools. While prior literature suggests green space is associated with positive health factors [39], the fact that this green space is nearly ubiquitous in neighborhoods near these schools may cause our model to down-weight these urban design factors.

The evidence of what our deep learning model found to be indicative of mortality can be corroborated with existing literature on associations of these image features with higher socioeconomic status, decreased obesity, and greater designs in urban planning [39–41]. For instance, we found that sidewalks, driveways, curvy roads, hiking trails and baseball fields, amongst other factors, were related to lower mortality; these factors were also uncovered from inspection of learned convolutional filters (Supplementary Figures 1-4).

The assigned importance given to image features such as baseball fields does not imply that these components are necessary for urban design but does further elucidate a suite of modifiable community factors to design interventions. For instance, accessibility to trails, while indicative of a county with high socioeconomic status[42,43], provides a convenient means to exercise but can also provide access to other portions of a community, allowing more social mobility and access to health care[44,45].

Despite our strong findings, there were some limitations to our study. The design of this study was cross-sectional. This precluded our ability to show temporality and that the built environment was causally associated with mortality. The images also were taken in 2018 while we assessed mortality in 2015. While we do not expect areas surrounding schools to change substantially, we cannot confirm the extent of land-use alterations between 2015 and 2018. We assessed county mortality rates using a small percentage of schools from the county, making generalizations to counties included in the study and counties not in our dataset. However, we were able to predict crude mortality rates accurately within a subsample of these counties. Higher correspondence between predicted and true mortality may be achieved by increasing the number of schools selected. In addition, we may be able to identify more ubiquitous points of interest other than schools from which to sample.

While we acknowledge prior literature documenting the potential for shadow effects to confound aerial imagery analysis, we also note that our model was able to pick up on key factors associated with mortality by utilizing learned filters with shape, color and intensity[46,47] (Supplementary Figures 1-4). Possible removal or augmentation of these shadows may cause the model to focus on other important characteristics pertinent to higher mortality prediction [46–48]. Finally, although we sampled counties randomly, our sampling scheme preferred populous counties, and had assumed that these counties would solely contain suburban and residential land-use patterns. However, even populous counties contained rural areas that may have obfuscated our ability to sample ubiquitous land-use regions for the mortality prediction potentially biasing the results. Recent deep learning works have focused on grasping health factors such as access to care in the rural setting[49] and could be employed in the context of mortality in the future.

The deep learning approaches used were able to achieve remarkable performance given technical challenges associated with sampling images over large geographic regions. Nor were covariate measures or temporality incorporated into the model. Different sampling techniques [50], feature aggregation measures, evaluating or producing higher resolution images [51] and direct estimation and segmentation of various land-use objects can potentially provide more accurate and interpretable models for studying health and disease [22,52,53]. The images may have also been sampled during various points throughout the year, thus results may have been affected by seasonality, however, it is likely that the images had been collected randomly with respect to seasonality and the fact that the model was able to distinguish mortality rates further attests of the ability to learn features less tied to seasonality. Future applications could include deep learning methodologies that are able to account for effects of seasonality[54,55]. Transfer learning from other GIS-corroborated data may also improve the model's performance. Regardless, our approach is scalable and uses open access data enabling further exploration.

This approach to assess mortality using CNNs has much external applicability. Offshoots of this approach may further explore how the built environment affects mortality in more precise estimates, such as cities, or explore rural areas exclusively[56]. Our design may particularly be beneficial in developing countries where access to covariates is limited and a deep learning modeling approach is more feasible and inexpensive. Applications beyond assessing mortality as an outcome could benefit from this approach and could use other landmarks as a sampling strategy for prediction instead of schools. Ultimately, these tools could then be used by epidemiologists and policy makers to identify clusters of exposures or diseases such as cancer, arsenic exposure, or infectious diseases for more targeted interventions[22].

## Conclusion

We found that mortality could be predicted from satellite imagery. Future use of deep learning and satellite imagery may assist in forming targeted public health interventions and policy changes.

**Author Contributions:**

    Concept and design: JJL, RML, and TAM

    Acquisition, analysis, or interpretation of the data: All authors

    Drafting of the manuscript: JJL, RML, and TAM

    Critical revision to the manuscript for important intellectual content: All authors

    Statistical analysis: JJL, RML, TAM

    Administrative, technical, or material support: JJL, RML and TAM

    Supervision: TAM

**Funding:** This work was supported by NIH grant R01CA216265. JL is supported through the Burroughs Wellcome Fund Big Data in the Life Sciences at Dartmouth. RML was funded under NIAID 2T32AI007519-21.

**Acknowledgements:** The authors would like to thank Christopher L. Henry and Jorge F. Lima for discussion of image interpretations.

**Conflicts of Interest:** The authors report no conflicts of interest.

**Competing Financial Interests:** The authors have no competing financial interests.

# Figures

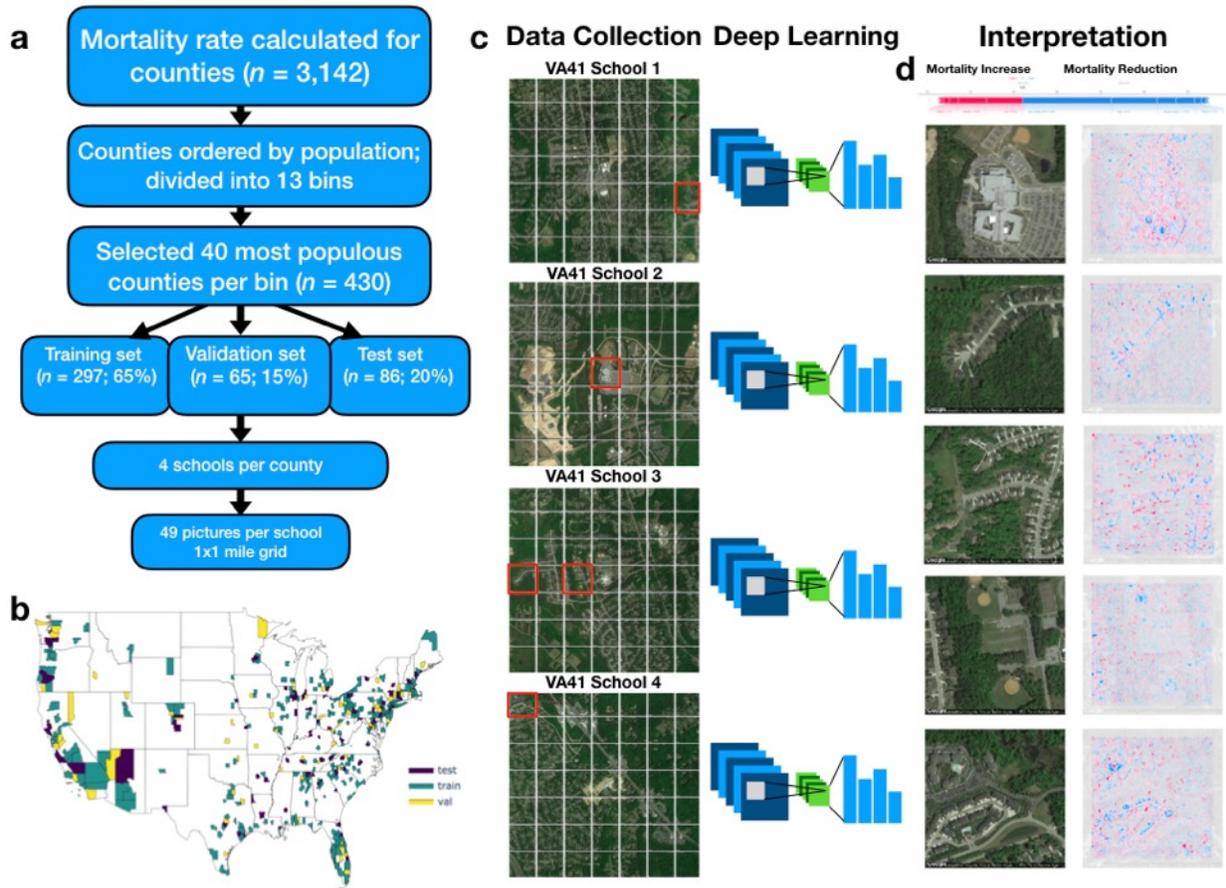

**Figure 1: Method Overview:** a) Flow diagram depicting selection of counties and sampling scheme for images; 13 mortality bins created from 1000 most populous counties to capture diversity of mortality rates; b) Map of the United States colored by selection of training, testing and validation counties; c) 4 sets of 1x1 mile grid of images (7 images by 7 images) were selected per each county, all images randomized and used to train deep learning model; visual depiction of deep learning: filters slide across each image and pick up on key image features, information is aggregated to form model predictions; the predictions for each of these 196 images per county are averaged to yield final mortality rates; d) image features interrogated using SHAP and compared to important covariates from linear model for each test county.

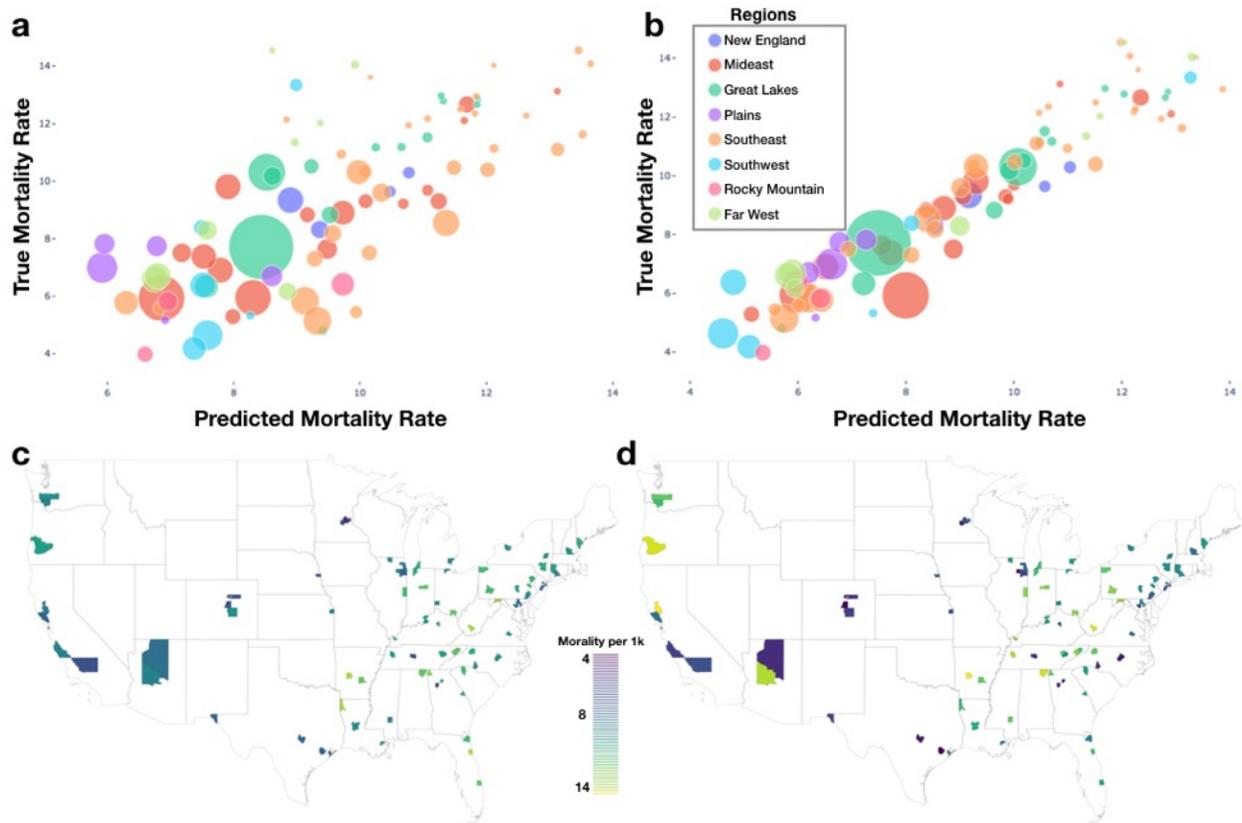

**Figure 2: Model Results:** a) deep learning predicted versus true mortality for each test county; b) linear model predicted versus true mortality for each test county; a-b) legend contains look-up dictionary for naming of US regions featured in the rest of the study; size of bubble is correspondent to population size of county; c) predicted mortality plotted geographically for test counties; d) true mortality plotted geographically for test counties

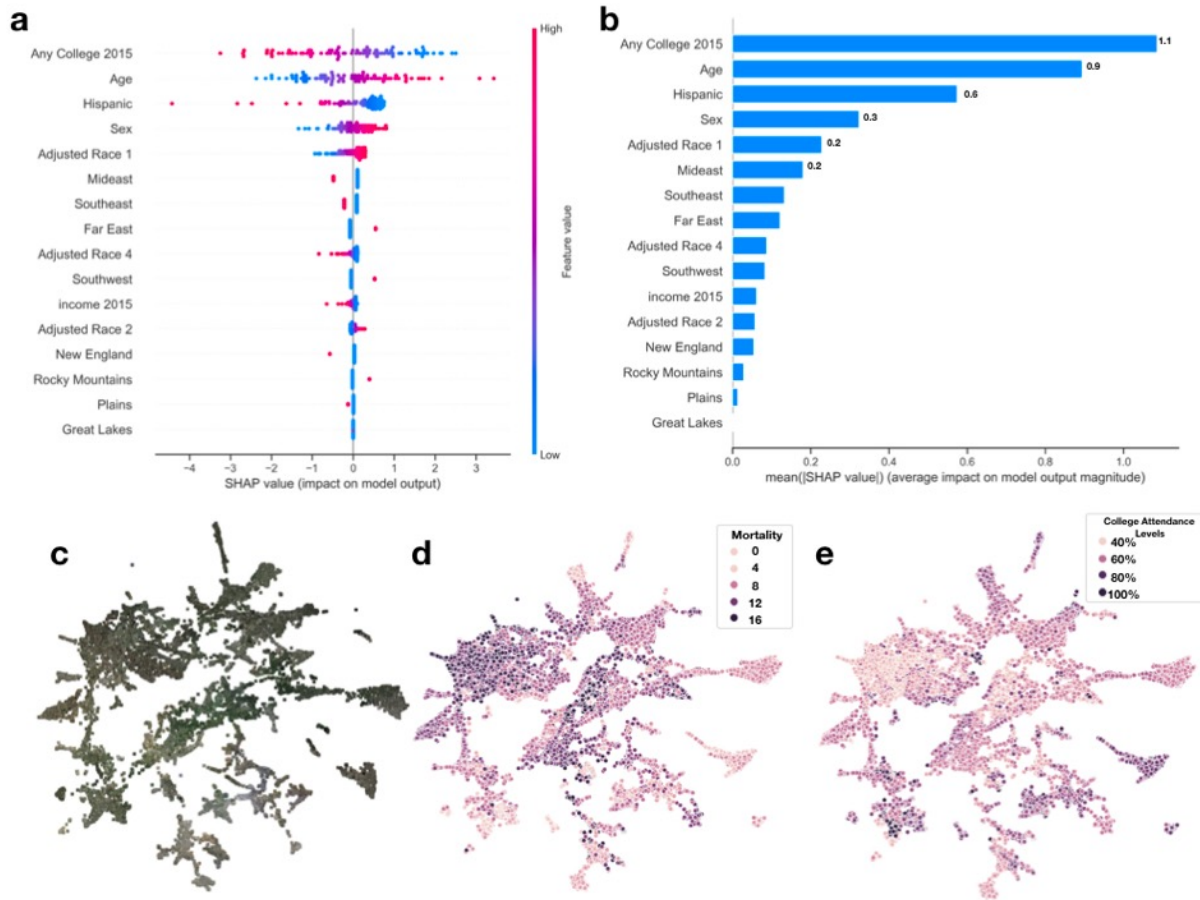

**Figure 3: SHAP Summary of Covariate Predictors and Image Embeddings** a) SHAP summary plot for covariate model evaluated on test set, covariates ranked by feature importance; b) SHAP rankings of overall importance of covariates; c) overall covariate importance over test set; d-e) 2-D UMAP plot of image features derived from images using deep learning model: d) actual images overlaying 2-D coordinates; e) colored by true county mortality; f) colored by average college attendance; we note here that our model has been explicitly trained to associate image features with mortality; the separability of college attendance over the image features is an artifact of the association between education status and mortality

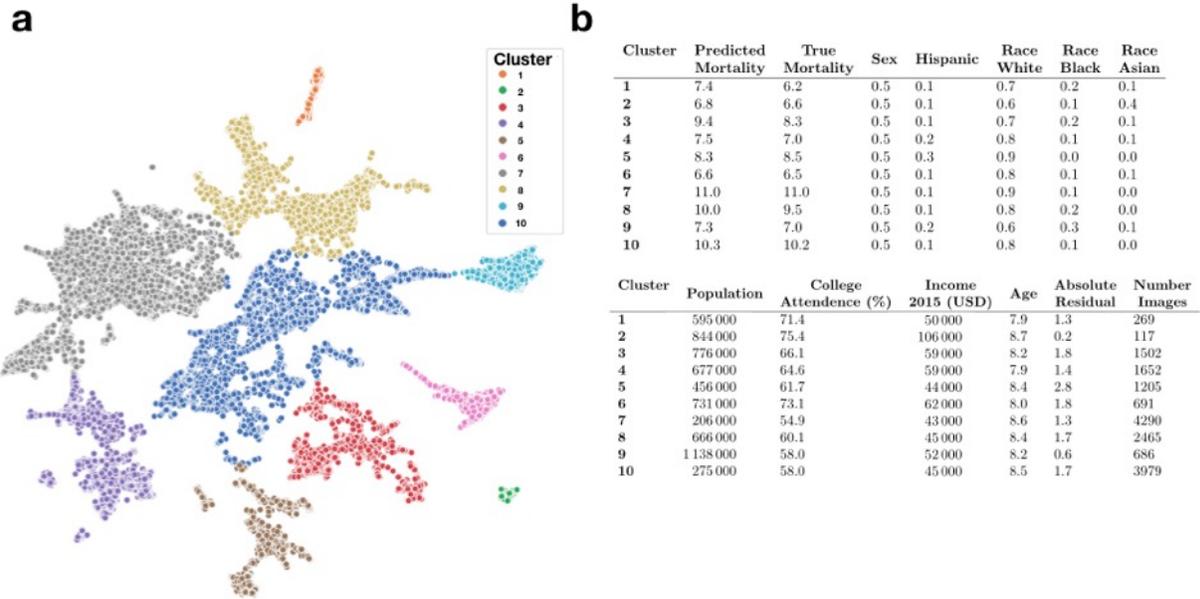

**Figure 4: Key Covariate Characteristics of Found Image Clusters:** a) 2-D UMAP plots colored by cluster; b) averaged covariate statistics for each image cluster; average population and income estimates rounded to thousands place

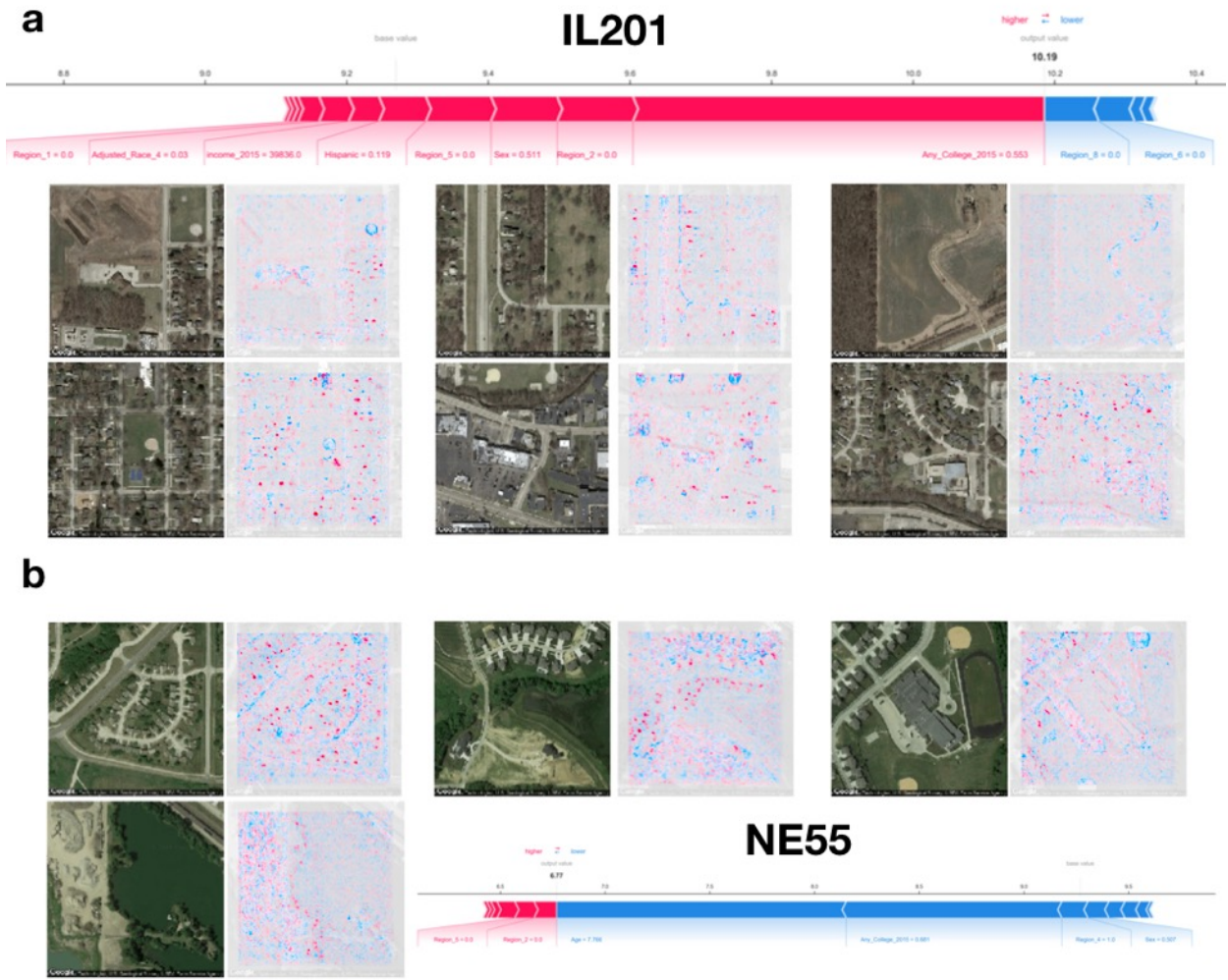

**Figure 5: SHAP Image Interpretations:** a-b) Shapley features extracted from both covariate and image models for select images from counties in Illinois and Nebraska (FIPS codes IL201 and NE55 respectively); blue coloring indicates features associated with reductions in mortality, red indicates association with increased mortality

# Supplementary Material

**Availability of High-Resolution Materials**
Due to large file size, high-resolution manuscript figures and ultra-high-resolution image of UMAP image embeddings overlaid with satellite images are available upon request.

**More Information on Selection of Variables for Covariate Model**
For adjusted linear models, age, sex, and Hispanic status, were extracted from the death certificates. Age was divided into 19 categories brackets and race was coded as White, Black, and Asian or Pacific Islander. Individuals considered Alaskan or Native American were removed prior to mortality calculations due to inconsistent coding of race between the CDC Wonder and SEER datasets. Percentage of college attainment was imputed for counties without information as the median value for all available countries. Median per capita county income and the median value for all available counties was also imputed for counties with any missing per capita income. Public school information was collected from the National Center for Education Statistics. Covariate information was collected from the SEER program database.

Ages were binned into 5-year brackets and thus turned into an ordinal variable. The deaths were averaged across the brackets via a weighted average to form an interpretation number of deaths. Race was changed into a ratio where the number of deaths of individuals in each county of a certain race were divided by the total number of individuals in each county creating a race ratio. Percentages of deaths were calculated by sex and Hispanic status. Median income and any college were not changed. Grouping regions include: New England, Mideast, Great Lakes, Plains, Southeast, Southwest, Rocky Mountain, and Far West.

|  | Training | Testing | Validation |
|---|---|---|---|
| n (%) | 279 (65) | 86 (20) | 65 (15) |
| *Mean (SD) for all counties* | | | |
| **Population (in 1000s)** | 537 (823) | 491 (677) | 505 (555) |
| **Mortality Rate (per 1000 people)** | 9.29 (2.8) | 9.21 (2.8) | 9.35 (2.9) |
| **Per capita income ($)\*** | 47000 (13000) | 48900 (17400) | 47500 (11800) |
| **Any College Rate (%)** | 60.0 (9.6) | 60.4 (10.4) | 60.8 (10.3) |

**Supplementary Table 1**: Average county characteristics by data set; 2 counties in the training set did not have income data, values were imputed by region; Averaged county mortality rates were not weighted by population size

**Supplementary Table 2**: Standardized Coefficients of Linear Regression Model

|  | coef | std err | t | P(T>|t|) | [0.025 | 0.975] |
|---|---|---|---|---|---|---|
| Constant | 9.39 | 0.05 | 192.32 | 0.00 | 9.29 | 9.48 |
| Age | 1.46 | 0.07 | 20.32 | 0.00 | 1.32 | 1.60 |
| Any College 2015 | −1.24 | 0.08 | −15.79 | 0.00 | −1.39 | −1.08 |
| Hispanic | −0.95 | 0.07 | −14.23 | 0.00 | −1.09 | −0.82 |
| Proportion Male | 0.43 | 0.06 | 6.67 | 0.00 | 0.30 | 0.56 |
| Far West | 0.30 | 0.05 | 5.67 | 0.00 | 0.20 | 0.41 |
| Southwest | 0.26 | 0.05 | 5.09 | 0.00 | 0.16 | 0.36 |
| Proportion Asian | −0.22 | 0.06 | −3.77 | 0.00 | −0.34 | −0.11 |
| Income 2015 | −0.20 | 0.08 | −2.64 | 0.01 | −0.35 | −0.05 |
| Mideast | −0.19 | 0.05 | −4.22 | 0.00 | −0.28 | −0.10 |
| Southeast | −0.15 | 0.04 | −3.28 | 0.00 | −0.23 | −0.06 |
| New England | −0.12 | 0.05 | −2.41 | 0.02 | −0.21 | −0.02 |
| Proportion White | 0.12 | 0.03 | 3.54 | 0.00 | 0.05 | 0.18 |
| Plains | −0.06 | 0.05 | −1.19 | 0.24 | −0.16 | 0.04 |
| Rocky Mountains | 0.02 | 0.05 | 0.47 | 0.64 | −0.08 | 0.13 |
| Proportion Black | −0.02 | 0.04 | −0.45 | 0.65 | −0.10 | 0.07 |
| Great Lakes | 0.01 | 0.04 | 0.17 | 0.86 | −0.08 | 0.09 |

**Supplementary Table 3**: Non-Standardized Coefficients of Linear Regression Model

|  | coef | std err | t | P(T>|t|) | [0.025 | 0.975] |
|---|---|---|---|---|---|---|
| Constant | −15.5 | 2.2 | −7.1 | 0.0 | −19.7 | −11.2 |
| Proportion White | −3.4 | 0.7 | −4.6 | 0.0 | −4.8 | −1.9 |
| Proportion Black | −4.3 | 1.0 | −4.5 | 0.0 | −6.2 | −2.5 |
| Proportion Asian | −7.8 | 1.0 | −7.7 | 0.0 | −9.7 | −5.8 |
| Hispanic | −6.6 | 0.5 | −14.2 | 0.0 | −7.5 | −5.7 |
| Proportion Male | 42.5 | 6.4 | 6.7 | 0.0 | 30.0 | 55.1 |
| Mean Age | 2.1 | 0.1 | 20.3 | 0.0 | 1.9 | 2.3 |
| Any College 2015 | −12.6 | 0.8 | −15.8 | 0.0 | −14.1 | −11.0 |
| New England | −2.5 | 0.4 | −6.7 | 0.0 | −3.3 | −1.8 |
| Mideast | −2.5 | 0.3 | −8.0 | 0.0 | −3.1 | −1.9 |
| Great Lakes | −2.0 | 0.3 | −6.3 | 0.0 | −2.6 | −1.4 |
| Plains | −2.2 | 0.3 | −6.8 | 0.0 | −2.9 | −1.6 |
| Southeast | −2.3 | 0.3 | −7.2 | 0.0 | −2.9 | −1.7 |
| Rocky Mountains | −1.8 | 0.3 | −5.4 | 0.0 | −2.5 | −1.2 |
| Far West | −1.1 | 0.3 | −3.7 | 0.0 | −1.6 | −0.5 |
| Southwest | −1.1 | 0.3 | −3.3 | 0.0 | −1.7 | −0.4 |
| Income 2015 | $-1.6 \cdot 10^{-05}$ | $5.9 \cdot 10^{-06}$ | −2.6 | 0.0 | $-2.7 \cdot 10^{-05}$ | $-4.0 \cdot 10^{-06}$ |

**Supplementary Table 4**: Fit Statistics for Linear Regression Model on Training and Validation Set

| R-squared: | 0.904 |
|---|---|
| Adj. R-squared: | 0.900 |
| F-statistic: | 221.2 |
| Prob (F-statistic): | 7.07E-158 |
| Log-Likelihood: | -446.16 |
| AIC: | 922.3 |
| BIC: | 979.9 |

**Supplementary Table 5**: SHAP Feature Importance Scores over Test Set

| Any College 2015 | 1.1 |
| Mean Age | 0.9 |
| Hispanic | 0.6 |
| Proportion Male | 0.3 |
| Proportion White | 0.2 |
| Mideast | 0.2 |
| Proportion Black | 0.1 |
| Proportion Asian | 0.1 |
| Income 2015 | 0.1 |
| New England | 0.1 |
| Southeast | 0.1 |
| Southwest | 0.1 |
| Far West | 0.1 |
| Great Lakes | 0.0 |
| Plains | 0.0 |
| Rocky Mountains | 0.0 |

**Supplementary Table 6**: Weighted T-Test Comparison between Clusters for Differences in Key Covariates: a) Mortality, b) Income, c) Hispanic, d) Any College, e) Age:

| a | 1 | 2 | 3 | 4 | 5 | 6 | 7 | 8 | 9 | 10 |
|---|---|---|---|---|---|---|---|---|---|---|
| 1 | | 2E-01 | 1E-40 | 3E-07 | 1E-21 | 3E-01 | 4E-194 | 4E-103 | 3E-20 | 1E-121 |
| 2 | | | 2E-13 | 5E-01 | 4E-07 | 1E+00 | 5E-78 | 2E-39 | 7E-07 | 2E-46 |
| 3 | | | | 2E-50 | 1E+00 | 2E-75 | 6E-278 | 1E-58 | 3E-43 | 1E-126 |
| 4 | | | | | 5E-37 | 5E-09 | 0E+00 | 7E-228 | 1E+00 | 0E+00 |
| 5 | | | | | | 4E-42 | 4E-167 | 3E-26 | 1E-24 | 2E-70 |
| 6 | | | | | | | 0E+00 | 2E-207 | 2E-18 | 8E-256 |
| 7 | | | | | | | | 8E-125 | 0E+00 | 9E-51 |
| 8 | | | | | | | | | 4E-156 | 1E-21 |
| 9 | | | | | | | | | | 2E-198 |
| 10 | | | | | | | | | | |

| b | 1 | 2 | 3 | 4 | 5 | 6 | 7 | 8 | 9 | 10 |
|---|---|---|---|---|---|---|---|---|---|---|
| 1 | | 1E-235 | 2E-05 | 4E-05 | 2E-24 | 1E-25 | 4E-32 | 2E-16 | 4E-05 | 4E-12 |
| 2 | | | 3E-69 | 1E-69 | 0E+00 | 1E-117 | 0E+00 | 0E+00 | 0E+00 | 0E+00 |
| 3 | | | | 1E+00 | 1E-64 | 2E-01 | 1E-210 | 6E-101 | 5E-08 | 2E-138 |
| 4 | | | | | 1E-63 | 2E-01 | 1E-211 | 1E-99 | 4E-07 | 4E-139 |

|   | 1 | 2 | 3 | 4 | 5 | 6 | 7 | 8 | 9 | 10 |
|---|---|---|---|---|---|---|---|---|---|---|
| 5 |   |   |   |   |   | 1E-157 | 6E-04 | 9E-03 | 9E-85 | 1E+00 |
| 6 |   |   |   |   |   |   | 0E+00 | 3E-213 | 1E-40 | 2E-206 |
| 7 |   |   |   |   |   |   |   | 4E-26 | 1E-118 | 1E-18 |
| 8 |   |   |   |   |   |   |   |   | 5E-71 | 1E+00 |
| 9 |   |   |   |   |   |   |   |   |   | 5E-54 |
| 10 |   |   |   |   |   |   |   |   |   |   |

| c | 1 | 2 | 3 | 4 | 5 | 6 | 7 | 8 | 9 | 10 |
|---|---|---|---|---|---|---|---|---|---|---|
| 1 |   | 2E-31 | 4E-01 | 2E-17 | 4E-47 | 1E+00 | 1E-147 | 3E-06 | 1E-274 | 1E-03 |
| 2 |   |   | 5E-09 | 7E-03 | 1E-15 | 3E-05 | 1E-171 | 2E-20 | 2E-101 | 3E-12 |
| 3 |   |   |   | 1E-90 | 4E-223 | 1E+00 | 9E-243 | 2E-07 | 1E-184 | 2E-05 |
| 4 |   |   |   |   | 1E-75 | 8E-43 | 0E+00 | 2E-179 | 1E+00 | 2E-217 |
| 5 |   |   |   |   |   | 5E-110 | 0E+00 | 0E+00 | 5E-49 | 0E+00 |
| 6 |   |   |   |   |   |   | 7E-149 | 2E-05 | 7E-109 | 8E-04 |
| 7 |   |   |   |   |   |   |   | 8E-207 | 0E+00 | 2E-174 |
| 8 |   |   |   |   |   |   |   |   | 1E-319 | 1E+00 |
| 9 |   |   |   |   |   |   |   |   |   | 1E-221 |
| 10 |   |   |   |   |   |   |   |   |   |   |

| d | 1 | 2 | 3 | 4 | 5 | 6 | 7 | 8 | 9 | 10 |
|---|---|---|---|---|---|---|---|---|---|---|
| 1 |   | 5E-13 | 7E-17 | 2E-17 | 8E-64 | 4E-06 | 3E-139 | 5E-100 | 8E-259 | 1E-116 |
| 2 |   |   | 4E-23 | 2E-19 | 4E-59 | 8E-07 | 1E-96 | 1E-83 | 3E-291 | 3E-88 |
| 3 |   |   |   | 6E-03 | 7E-33 | 1E-69 | 2E-264 | 4E-90 | 5E-90 | 4E-167 |
| 4 |   |   |   |   | 2E-11 | 1E-65 | 2E-191 | 6E-43 | 2E-40 | 5E-103 |
| 5 |   |   |   |   |   | 1E-188 | 5E-91 | 2E-07 | 6E-25 | 7E-33 |
| 6 |   |   |   |   |   |   | 0E+00 | 3E-290 | 0E+00 | 0E+00 |
| 7 |   |   |   |   |   |   |   | 1E-96 | 7E-14 | 1E-44 |
| 8 |   |   |   |   |   |   |   |   | 5E-10 | 3E-19 |

| e | 1 | 2 | 3 | 4 | 5 | 6 | 7 | 8 | 9 | 10 |
|---|---|---|---|---|---|---|---|---|---|---|
| 9 | | | | | | | | | | 1E+00 |
| 10 | | | | | | | | | | |

| e | 1 | 2 | 3 | 4 | 5 | 6 | 7 | 8 | 9 | 10 |
|---|---|---|---|---|---|---|---|---|---|---|
| 1 | | 2E-104 | 2E-31 | 1E+00 | 2E-14 | 1E+00 | 7E-153 | 8E-43 | 1E-41 | 1E-89 |
| 2 | | | 2E-30 | 2E-87 | 3E-02 | 1E-157 | 1E+00 | 7E-05 | 5E-102 | 2E-04 |
| 3 | | | | 6E-113 | 3E-08 | 5E-60 | 2E-218 | 6E-29 | 4E-04 | 5E-84 |
| 4 | | | | | 7E-78 | 2E-03 | 0E+00 | 2E-208 | 2E-56 | 0E+00 |
| 5 | | | | | | 3E-31 | 2E-37 | 1E+00 | 7E-09 | 4E-05 |
| 6 | | | | | | | 0E+00 | 5E-92 | 3E-60 | 2E-192 |
| 7 | | | | | | | | 2E-69 | 3E-172 | 8E-56 |
| 8 | | | | | | | | | 5E-30 | 2E-05 |
| 9 | | | | | | | | | | 4E-77 |
| 10 | | | | | | | | | | |

**Supplementary Table 7**: Goodness-of-fit for univariable regression of predicted mortality to each of the mortality covariate predictors to directly associate image features with mortality factors

| Covariate | Adjusted R-Squared | P-Value |
|---|---|---|
| **Mean Age** | 0.971 | **4.35E-67** |
| **Proportion Male** | 0.961 | **8.55E-62** |
| **Proportion White** | 0.944 | **4.33E-55** |
| **Any College 2015** | 0.894 | **1.85E-43** |
| **Income 2015** | 0.797 | **2.29E-31** |
| **Proportion Black** | 0.513 | **3.92E-15** |
| **Southeast** | 0.402 | **2.62E-11** |
| **Hispanic** | 0.341 | **1.78E-09** |
| **Proportion Asian** | 0.293 | **3.65E-08** |
| **Mideast** | 0.177 | **2.95E-05** |
| **Great Lakes** | 0.160 | **7.51E-05** |
| **Far West** | 0.057 | **1.47E-02** |
| **New England** | 0.036 | **4.26E-02** |
| **Southwest** | 0.034 | **4.81E-02** |
| **Plains** | 0.017 | 1.19E-01 |
| **Rocky Mountains** | 0.010 | 1.7E-01 |

**Supplementary Table 8**: Further Description/Lookup Dictionary of Covariates included in Linear Model

| Model Covariate Name | Description |
|---|---|
| Region 1 | New England |
| Region 2 | Mideast |
| Region 3 | Great Lakes |
| Region 4 | Plains |
| Region 5 | Southeast |
| Region 6 | Southwest |
| Region 7 | Rocky Mountains |
| Region 8 | Far West |
| Sex | 0=Female, 1=Male |
| Age=0 | 0 years |
| Age=1 | 1-4 years |
| Age=17 | 80-84 years |
| Age=18 | 85+ years |
| Hispanic | 0=Non-Hispanic, 1=Hispanic |
| Adjusted Race | 1=White, 2=Black, 4=Asian/Pacific Islander |

**Further Description of Found SHAP and Convolutional Filter Features**

Shadows under buildings, trees, and open highway space were associated with increased mortality, which could be an artifact of higher mortality rates in rural counties and possibly higher instances of foliage within these counties may produce more shadows. We found that the color contrast between the center and borders of roofs was associated with increased mortality. We hypothesize that lack of maintenance or longevity of the building can be attributed to aging or socioeconomic indicators which are also independently associated with mortality. Further inspection of learned image filters demonstrated the ability of our model to identify these factors by utilizing shape and light intensity.

**Supplementary Figure 1:** Learned Convolutional Filters, numbered; we note the recurrence of patterns with parallel lines in different rotational orientations (filters 5, 8, 18, 22, 28, 50, 51, 54, 62, 64); similarities between filters 2, 6, 20; similarities between filters 15 and 42. Some filters place more weight on shape and morphology, while others focus more on color and intensity, as reflected in Supplementary Figures 2-4

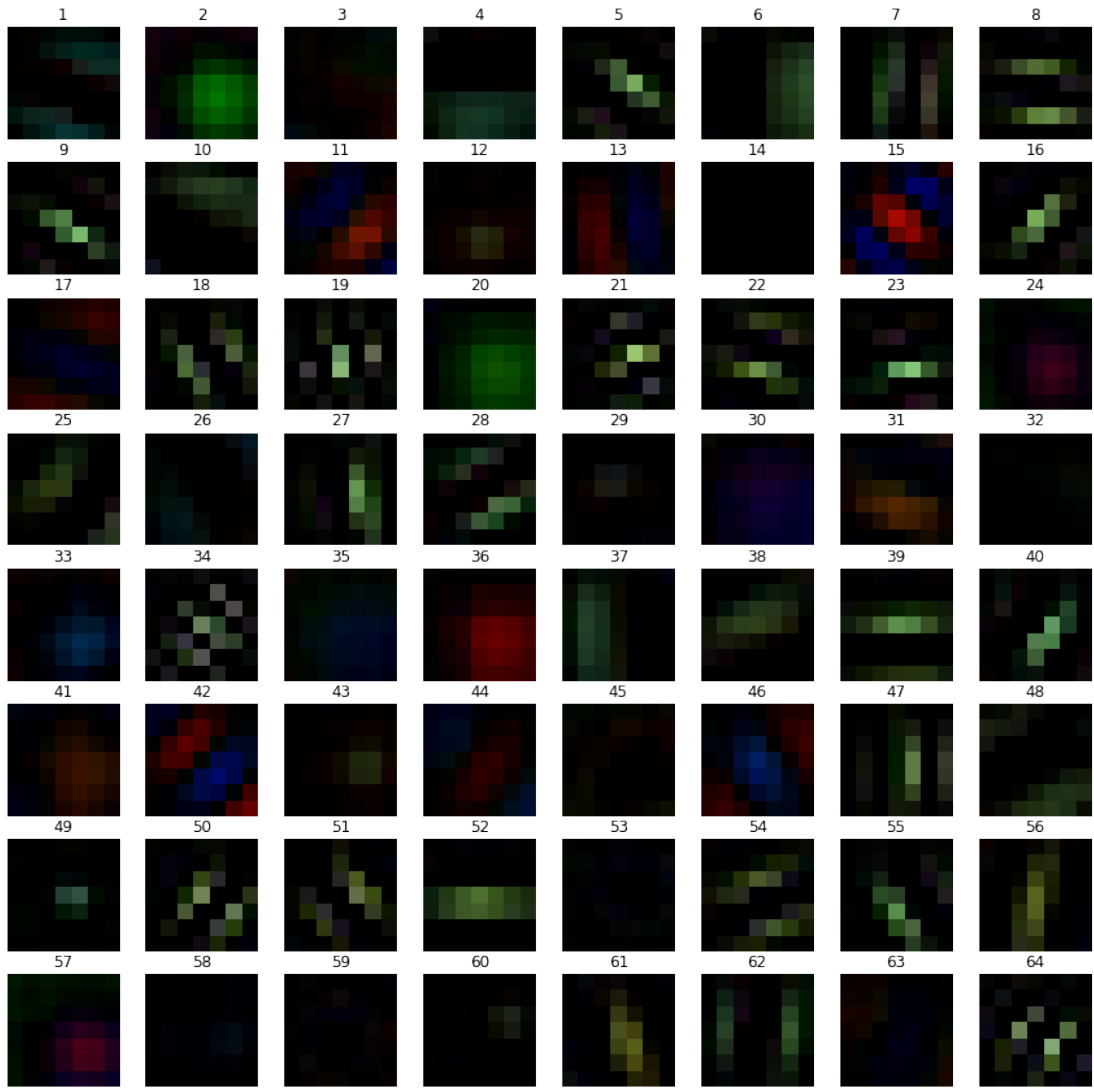

**Supplementary Figure 2**: Convolutional Filter Activations from VA41, Sampled Image around School 1; filter numbers correspond to the filters introduced in Supplementary Figure 1

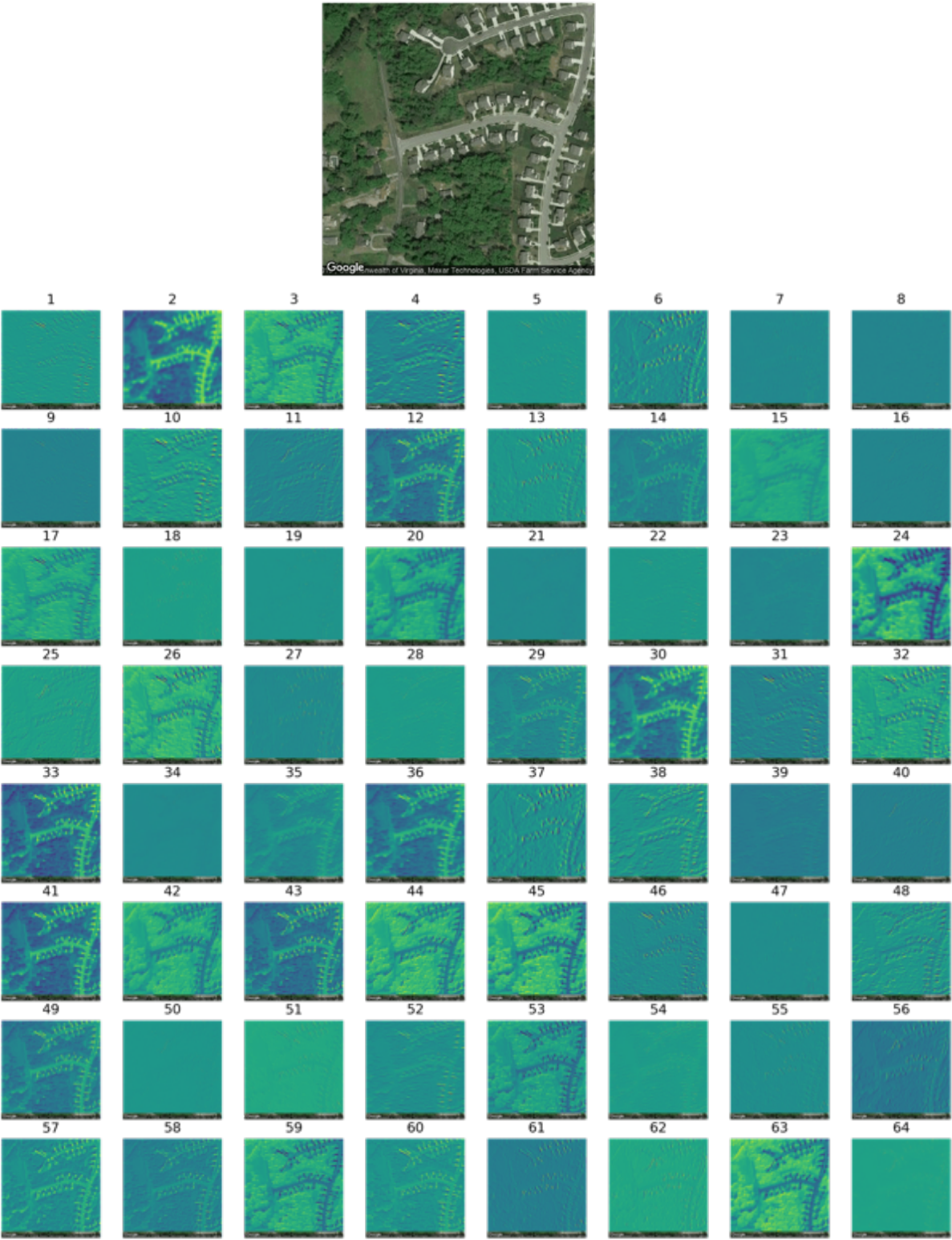

**Supplementary Figure 3:** Convolutional Filter Activations from CA75, Sampled Image around School 2; filter numbers correspond to the filters introduced in Supplementary Figure 1

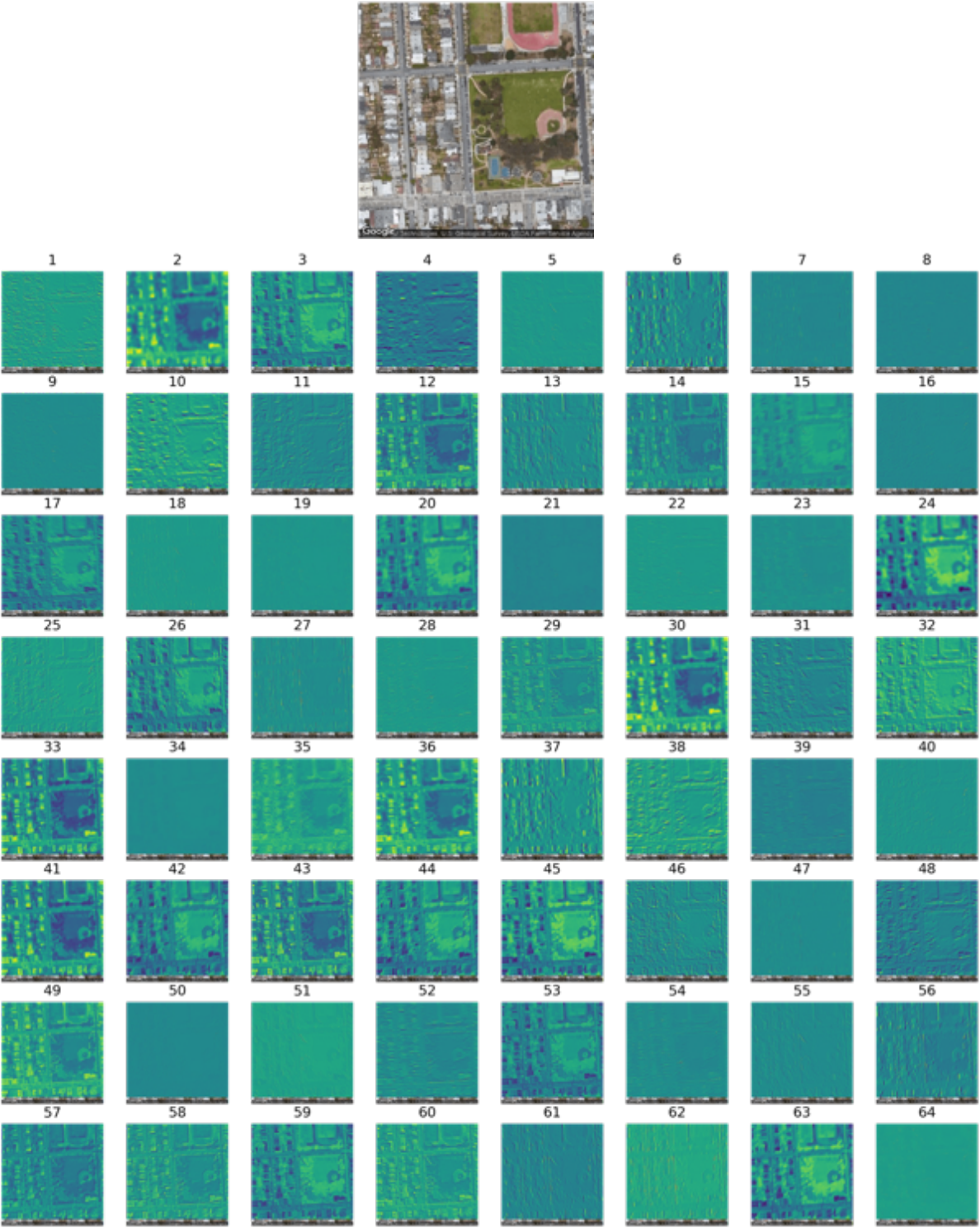

**Supplementary Figure 4**: Convolutional Filter Activations from CA75, Sampled Image around School 4; filter numbers correspond to the filters introduced in Supplementary Figure 1

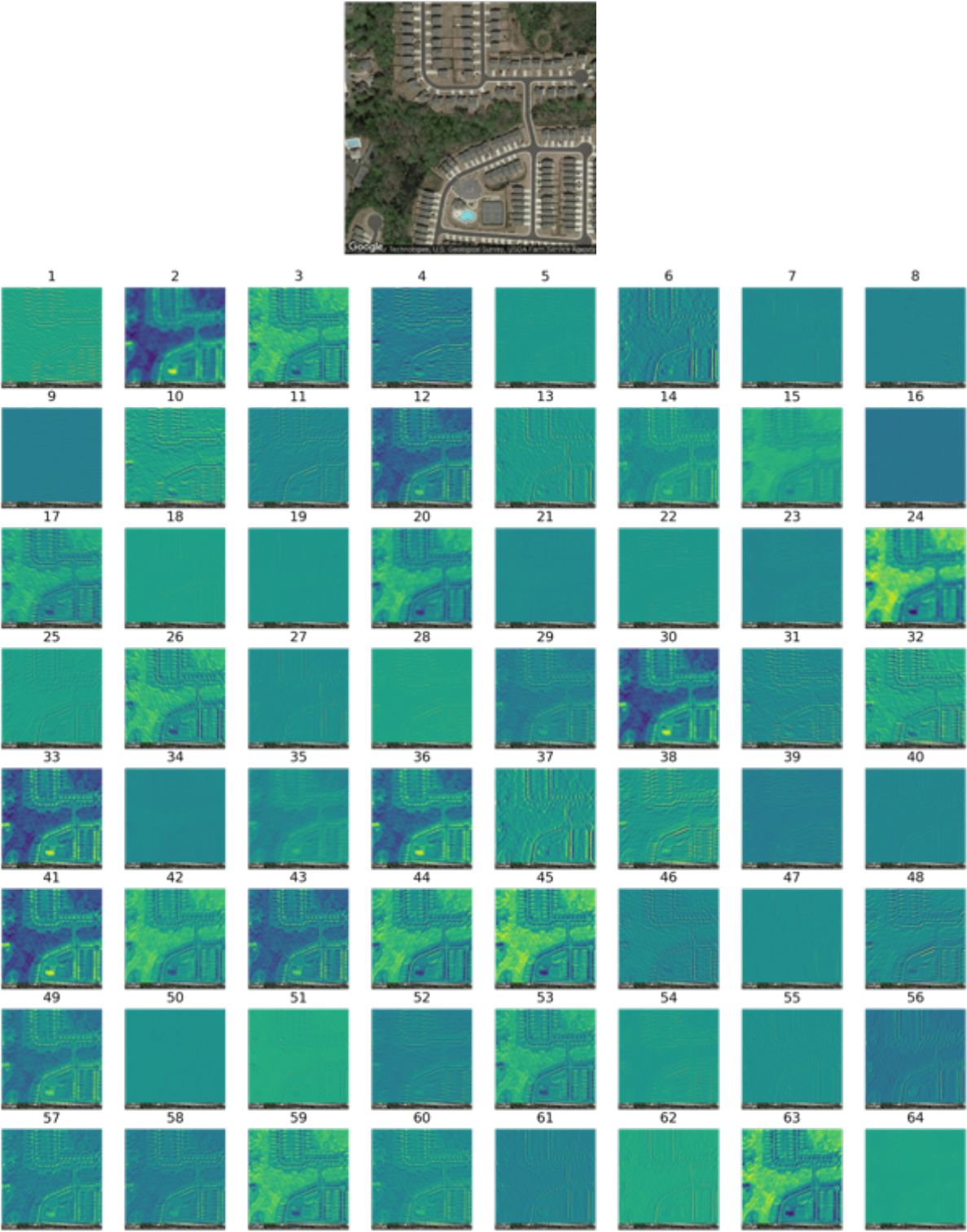

**Supplementary Figure 5:** Decomposition of Each Image Cluster by Region in US

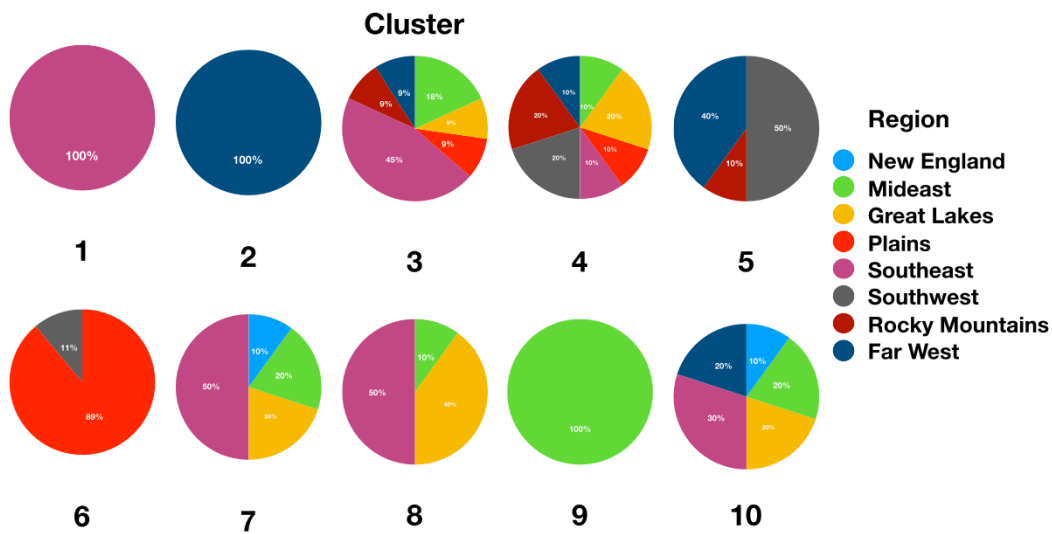

**Further Description of Image Plot Embeddings**

We found that counties from certain regions tended to group together in the embedding plots, formed by 1000-d vectors extracted from each image using the neural network. We hypothesize that this is due to similar flora, landscaping practices, or similar distribution of schools throughout communities within the same region in the United States. The clustered image features demonstrated that satellite images across large geographic distances could also be related to other mortality predictors. For instance, we hypothesize that the images found in cluster 2 were associated with more affluent and well-educated suburban neighborhoods, given a high proportion of sidewalks and curved street design. Most of these images were taken from San Francisco County in California, where the average income and educational attainment were much greater than the averages found across the other clusters. One possible interpretation of the stark contrast in mortality rates between clusters 2 and 7 is that cluster 7 may have contained more images from a rural context. This is supported by the fact that the average county population significantly lower than all of the other clusters.

With regards to other mortality predictors, we noticed that cluster 5 image features, containing El Paso County, were associated with a large Hispanic population. Cluster 5 represented a county with moderate mortality, income and education, and primarily located in the southwestern and western United States. Of note here was a sub-cluster with a high Hispanic population, mostly comprised of images from El Paso County, Texas. The images within this sub-cluster were associated with lower mortality as compared to the rest of the image cluster.

**Supplementary Figure 6:** UMAP Embeddings of Cluster 7 Overlaid with Original Images

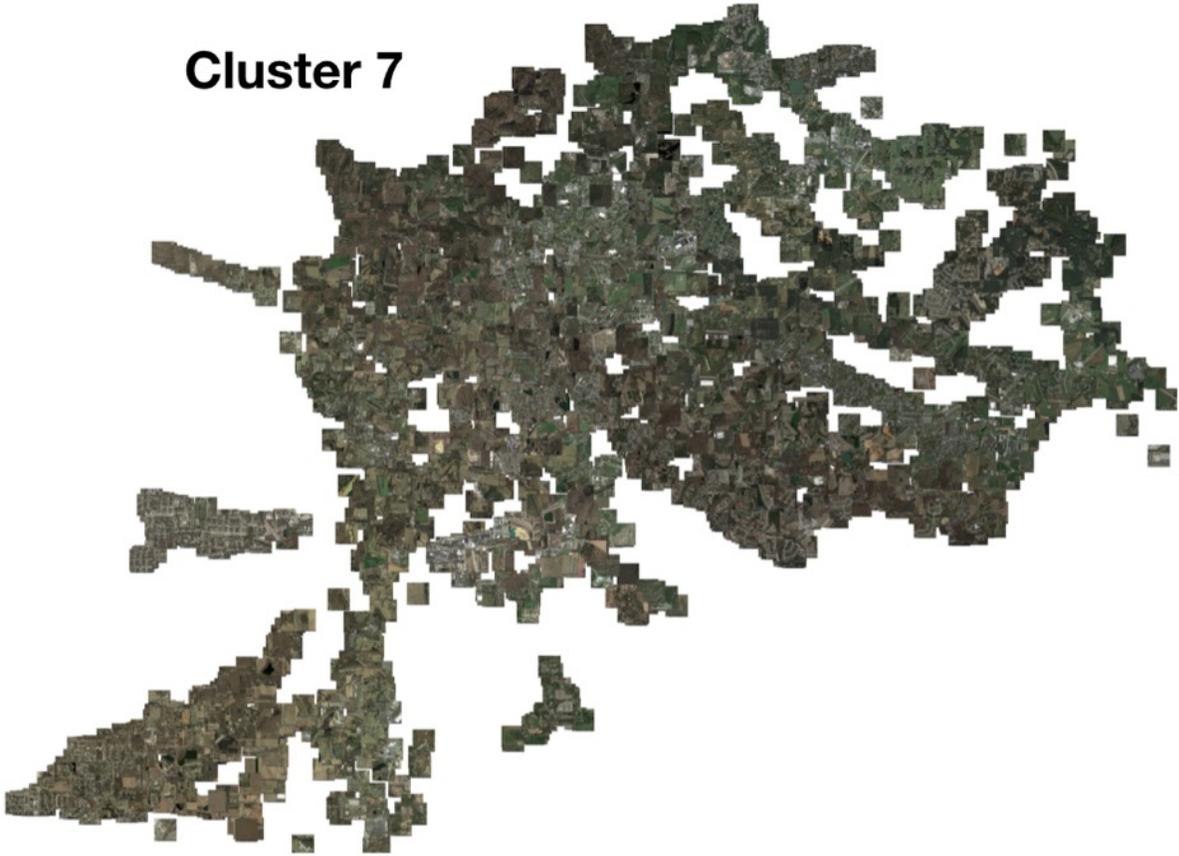

**Supplementary Figure 7:** UMAP Embeddings of Cluster 2 Overlaid with Original Images

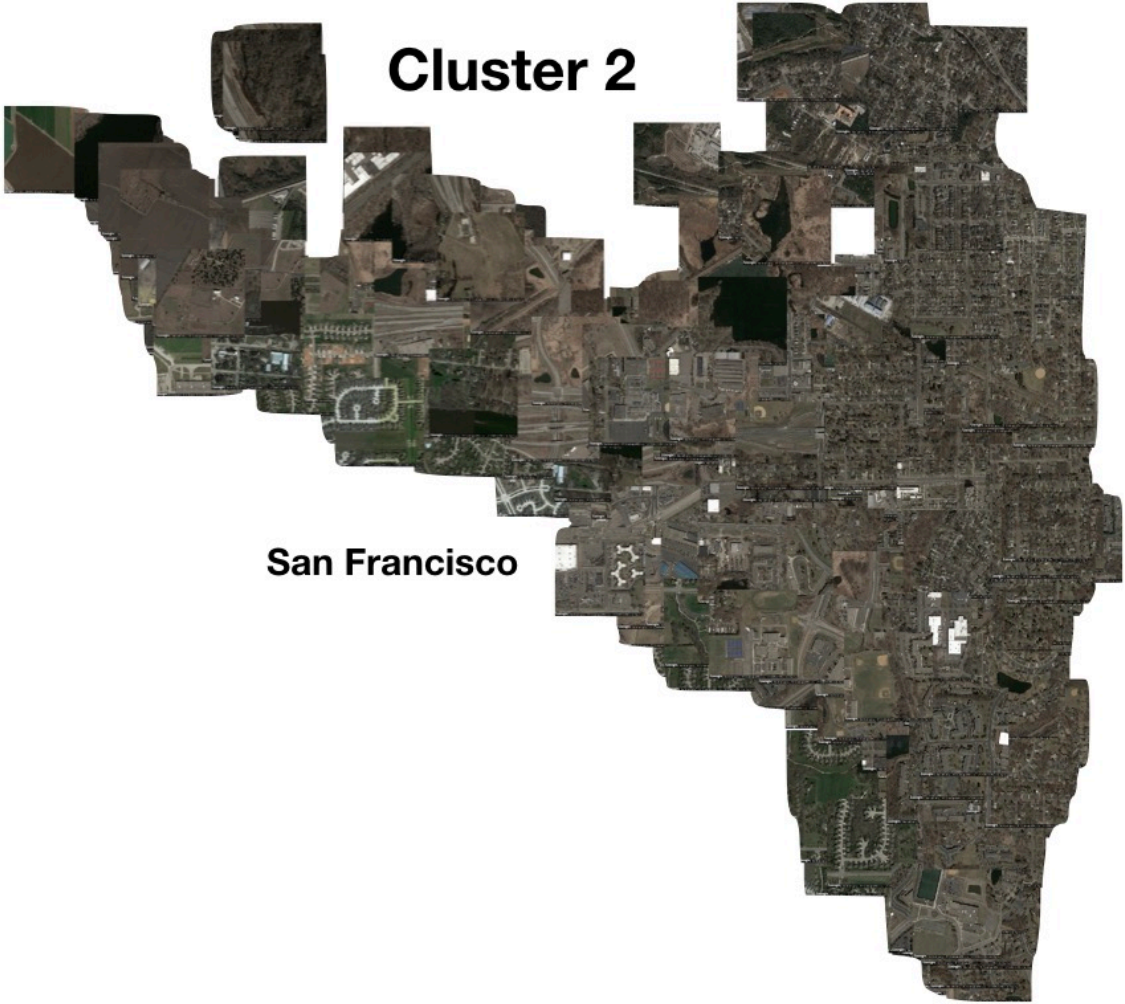

**Supplementary Figure 8**: UMAP Embeddings of Cluster 5 Overlaid with Original Images

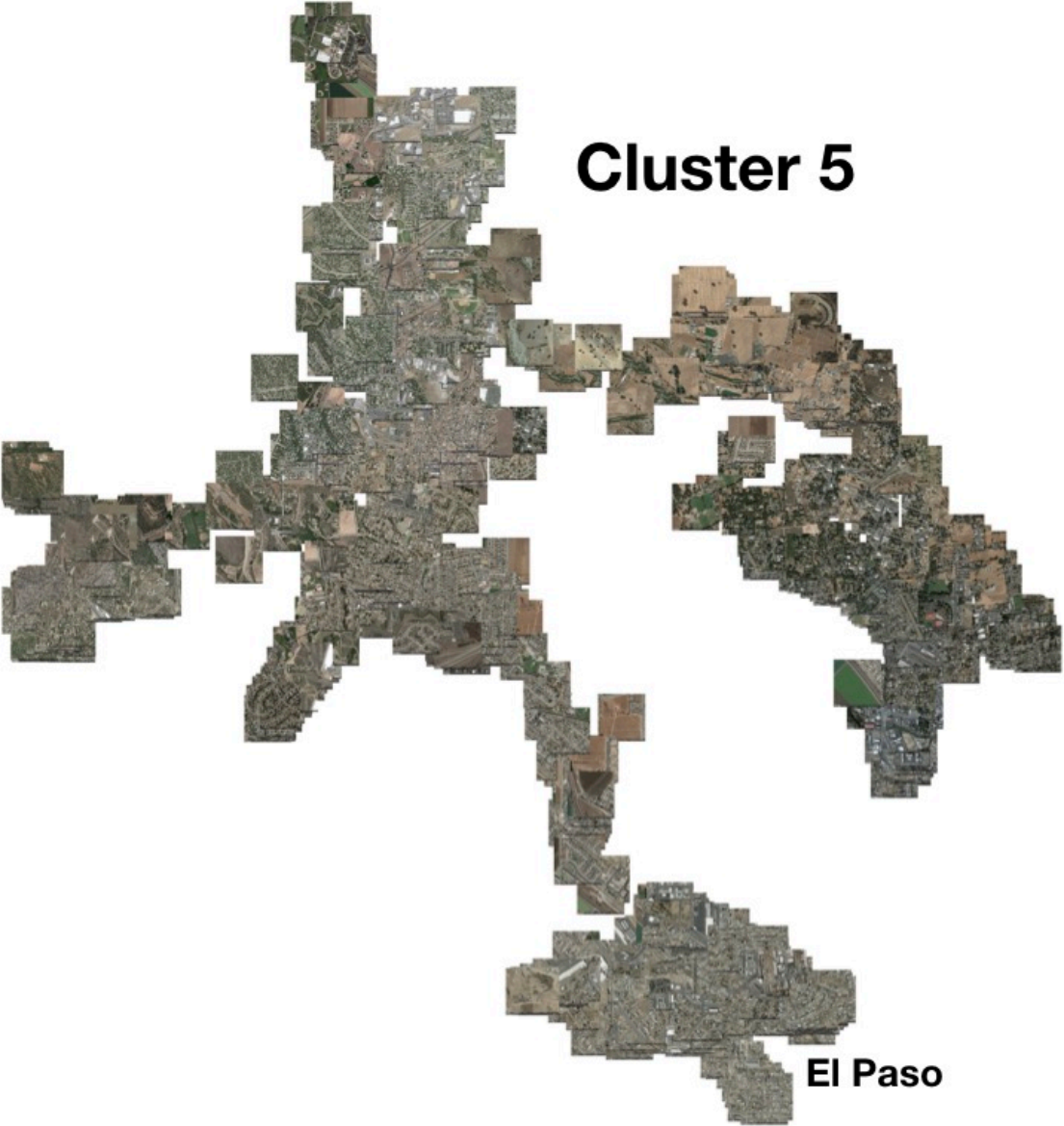